\pdfoutput=1
\documentclass[11pt]{article}

\usepackage[preprint]{acl}

\usepackage{times}
\usepackage{latexsym}
\usepackage{tcolorbox}
\usepackage{algorithm}
\usepackage[algo2e,ruled,linesnumbered,noline,noend]{algorithm2e}
\usepackage[T1]{fontenc}
\usepackage[utf8]{inputenc}

\usepackage{microtype}

\usepackage{inconsolata}

\usepackage{graphicx}

\usepackage{hyperref}
\usepackage{float}
\usepackage{booktabs}
\usepackage{multirow}
\usepackage{wrapfig}
\usepackage{natbib}
\usepackage{microtype}
\usepackage{xcolor}         % colors
\usepackage{amsmath}
\usepackage{amssymb}
\usepackage{multicol}
\usepackage{wrapfig}
\usepackage{tabularx}
\usepackage{booktabs}       % professional-quality tables
\usepackage{amsfonts}       % blackboard math symbols
\usepackage{graphicx}
\usepackage{nicefrac}  
\usepackage{pifont}
\usepackage{colortbl}
\usepackage{longtable}
\usepackage{ulem}
\usepackage{listings}
\usepackage{xspace}
\usepackage{fancyvrb}
\usepackage{adjustbox}
\usepackage{amssymb}
\usepackage{pifont}
\newcommand{\cmark}{\ding{51}}
\newcommand{\xmark}{\ding{55}}

\usepackage{amsthm}
\newtheorem{definition}{Definition}[section]  % 按section编号

\usepackage{tikz}
\usepackage{forest}

\definecolor{lightyellow}{RGB}{255,243,205}
\definecolor{lightblue}{RGB}{173,216,230}
\definecolor{lightgreen}{RGB}{144,238,144}

\title{A Survey on Personalized Alignment---The Missing Piece for Large Language Models in Real-World Applications}

\author{Jian Guan$^{1}$\thanks{Equal Contribution. $^\dagger$\text{Corresponding Author.}}, Junfei Wu$^{12*}$, Jia-Nan Li$^{13*}$, Chuanqi Cheng$^{13*}$, Wei Wu$^{1\dagger}$\\
$^1$Ant Group; $^2$Institute of Automation, Chinese Academy of Sciences; \\$^3$Gaoling School of Artificial Intelligence, Renmin University of China, Beijing, China.\\
\texttt{\{jianguanthu, wuwei19850318\}@gmail.com, junfei.wu@cripac.ia.ac.cn, }\\\texttt{\{lijianan,chengchuanqi\}@ruc.edu.cn}}

\begin{document}
\maketitle
\begin{abstract}
% Large Language Models (LLMs) have demonstrated remarkable capabilities, yet their transition to real-world applications reveals a critical limitation: the inability to adapt to individual preferences while maintaining alignment with universal human values. Current alignment techniques optimize for general helpfulness, honesty, and harmlessness through a one-size-fits-all approach, failing to accommodate users' diverse backgrounds and needs. This paper presents the first comprehensive survey of personalized alignment—a paradigm that enables LLMs to adapt their behavior within ethical boundaries based on individual preferences. We propose a unified framework comprising preference memory management, personalized generation, and feedback learning, analyzing existing techniques across prompting-based, encoding-based, parameter-based, and agent-based implementations. We examine evaluation protocols, real-world applications, and potential risks from individual privacy to societal polarization. Through systematic analysis of current approaches and future challenges, this survey provides a structured foundation for developing LLMs that better serve diverse user needs while upholding ethical principles.

Large Language Models (LLMs) have demonstrated remarkable capabilities, yet their transition to real-world applications reveals a critical limitation: the inability to adapt to individual preferences while maintaining alignment with universal human values. Current alignment techniques adopt a one-size-fits-all approach that fails to accommodate users' diverse backgrounds and needs. This paper presents the first comprehensive survey of personalized alignment—a paradigm that enables LLMs to adapt their behavior within ethical boundaries based on individual preferences. We propose a unified framework comprising preference memory management, personalized generation, and feedback-based alignment, systematically analyzing implementation approaches and evaluating their effectiveness across various scenarios. By examining current techniques, potential risks, and future challenges, this survey provides a structured foundation for developing more adaptable and ethically-aligned LLMs.
\end{abstract}

\begin{table*}[t]
\centering
\caption{Limitations of current one-size-fits-all alignment approaches. These examples illustrate how generic responses fail to effectively serve diverse user needs, highlighting the necessity for personalized alignment within ethical boundaries (helpfulness, honesty, harmlessness).}
\label{tab:multi_dimension_examples}
\begin{adjustbox}{max width=\linewidth}
\begin{tabular}{cccc}
\toprule
\multirow{2}{*}{\textbf{Individuals}}& \multicolumn{3}{c}{\textbf{Universal Human Value Constrains}} \\
\cmidrule{2-4}
& \multicolumn{1}{c}{\textbf{Helpfulness}} & \multicolumn{1}{c}{\textbf{Honesty}} & \multicolumn{1}{c}{\textbf{Harmlessness}} \\
\hline\hline
\multicolumn{4}{c}{\textbf{Prompt: }\textit{``What's your view on genetic editing technology?''}}\\
\midrule
\multicolumn{4}{p{22cm}}{\textbf{Generic Response from GPT-4o:  }\textit{``Genetic editing is an emerging technology with both potential benefits and risks. It could help treat diseases but also raises ethical concerns. Current research shows promising results while still requiring careful consideration ...''}}\\
\midrule
% \multirow{4}{*}{\textcolor{red}{{Bioethicist}}} 
\multirow{4}{*}{\textcolor{red}{\textbf{{User A}}}} & \textcolor{red}{More technical details} & \textcolor{red}{More scientific data and uncertainty disclosure} & \textcolor{red}{More detailed risk analysis} \\
\cmidrule{2-4}
& \multicolumn{3}{p{20cm}}{\textbf{Preferred Response:} ``Based on current clinical data, CRISPR gene editing shows 85\% efficacy in treating monogenic disorders \textcolor{red}{(technical details, scientific data)}, though uncertainty remains about off-target effects \textcolor{red}{(uncertainty disclosure)}. We must carefully weigh both therapeutic potential and long-term implications for human germline modification \textcolor{red}{(detailed risk analysis)} ....''} \\
\cmidrule{1-4}
% \multirow{4}{*}{\textcolor{blue}{{General Public}}} 
\multirow{4}{*}{\textcolor{blue}{\textbf{User B}}} & \textcolor{blue}{More practical implications} & \textcolor{blue}{More accessible facts} & \textcolor{blue}{Simpler risk explanation} \\
\cmidrule{2-4}
& \multicolumn{3}{p{20cm}}{\textbf{Preferred Response:} ``Think of genetic editing like a smart eraser that can fix mistakes in our DNA recipe \textcolor{blue}{(practical implications, accessible facts)}. While doctors have used it to help some patients with inherited diseases, they're still doing careful testing to make sure it's completely safe before wider use \textcolor{blue}{(simple risk explanation)} ...''} \\
\hline
\hline
\multicolumn{4}{c}{\textbf{Prompt: }\textit{``What happens after death?''}}\\\midrule
\multicolumn{4}{p{22cm}}{\textbf{Generic Response from GPT-4o:  }\textit{``The question of what happens after death has been contemplated across cultures and times. Different perspectives include scientific explanations of biological processes, various religious beliefs about afterlife, and philosophical discussions ...''}}\\
\midrule
% \multirow{4}{*}{\textcolor{red}{{Non-religious Person}}}
\multirow{4}{*}{\textcolor{red}{\textbf{User C}}}
& \textcolor{red}{More scientific explanations} & \textcolor{red}{More empirical evidence} & \textcolor{red}{More fact-based discussion} \\

\cmidrule{2-4}
& \multicolumn{3}{p{20cm}}{\textbf{Preferred Response:} ``From a biological perspective, death is characterized by measurable cessation of brain activity and organ functions \textcolor{red}{(scientific explanations)}. Medical research has documented this process through extensive clinical observations \textcolor{red}{(empirical evidence)}, though the exact moment of consciousness loss remains an active area of neurological study \textcolor{red}{(fact-based discussion)} ...''} \\
\cmidrule{1-4}
% \multirow{4}{*}{\textcolor{blue}{{Christian Believer}}} 
\multirow{4}{*}{\textcolor{blue}{\textbf{User D}}} 
& \textcolor{blue}{More spiritual guidance} & \textcolor{blue}{More biblical teachings} & \textcolor{blue}{More faith-based comfort}\\
\cmidrule{2-4}
& \multicolumn{3}{p{20cm}}{\textbf{Preferred Response:} ``Scripture teaches us that through faith in Christ, death is not an end but a transition to eternal life \textcolor{blue}{(biblical teachings, spiritual guidance)}. As Jesus promised in John 14:2, He has prepared a place for believers in His Father's house, offering us hope and peace in this divine assurance \textcolor{blue}{(faith-based comfort)} ...''} \\
\bottomrule
\end{tabular}
\end{adjustbox}
\end{table*}

\section{Introduction}

Large Language Models (LLMs) have revolutionized natural language processing tasks~\cite{achiam2023gpt,team2024gemini,guo2025deepseek}, achieved by aligning their behaviors with human preferences~\cite{ouyang2022training, bai2022training}. While current alignment techniques optimize for universal human values such as helpfulness, honesty, and harmlessness~\cite{askell2021general}, their transition to real-world applications reveals a crucial limitation: the inability to adapt to diverse user needs~\cite{kirk2023the,kasirzadeh2023conversation}, leading to reduced satisfaction and systematic bias against minority groups~\cite{siththaranjan2024distributional}, as exemplified in Table~\ref{tab:multi_dimension_examples}. Recent advances in LLMs, such as GPT-4.5~\cite{openai2024gpt45}, demonstrate improved capabilities in understanding user intent and showing greater ``emotional intelligence'', yet personalization remains a fundamental challenge that requires systematic solutions.

\begin{figure}[!t]
  \centering
  \includegraphics[width=\linewidth]{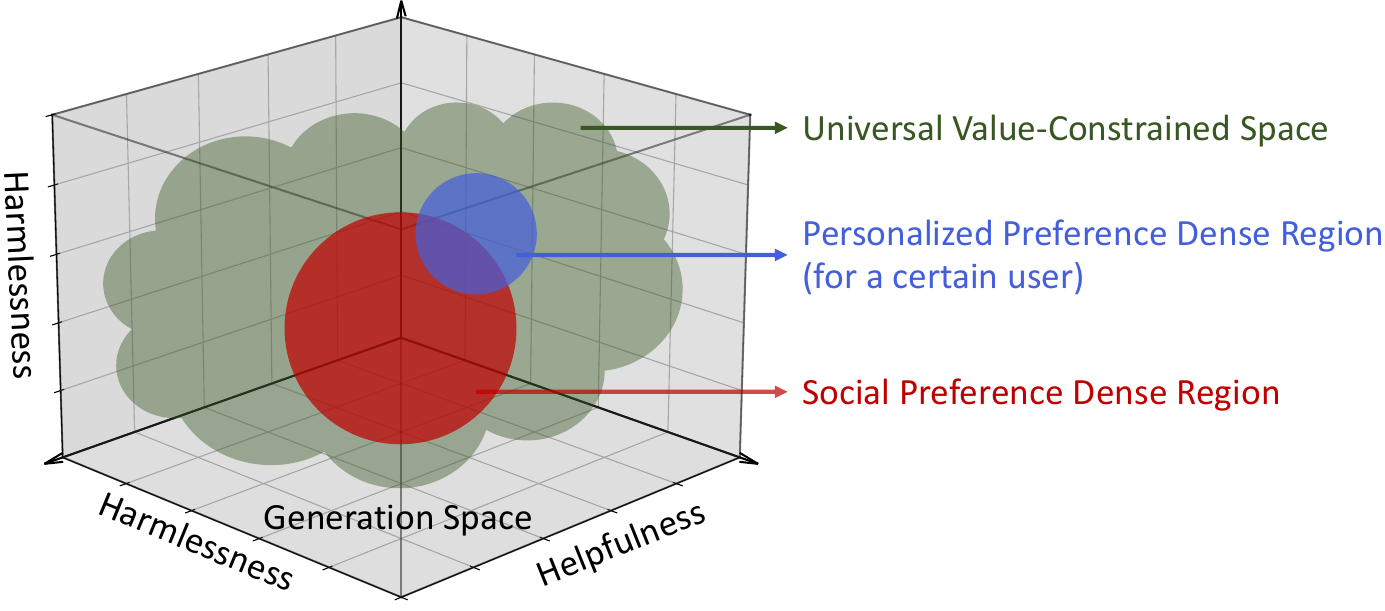}
  \caption{A visualization of the generation space for a certain prompt, illustrating the Pareto-optimal space of an LLM's responses under three dimensions of universal human values~\cite{rame2023rewarded}, with two distinct probability mass concentrations, where the social preference dense region emerges as the expected distribution across all personalized preference densities.}
  \label{fig:cube}
\end{figure}

Personalization has driven substantial socioeconomic value in traditional artificial intelligence (AI) systems like search engines~\cite{speretta2005personalized}, advertising~\cite{tucker2014social}, and recommendation~\cite{schafer2001commerce,guy2010social}. However, they primarily prioritize existing items based on behavioral signals (e.g., clicks, purchases). In contrast, LLM personalization must interpret a broader spectrum of preference indicators such as dialogue history~\cite{zhang2024personalization} and generate responses through sophisticated reasoning.
% present distinct challenges

To address the challenges, we propose a comprehensive personalized alignment framework with three components: (1) preference memory management for maintaining user-specific knowledge, (2) personalized generation and rewarding for incorporating personalized patterns, and (3) alignment through feedback for improving preference memory and generations. As illustrated in Figure~\ref{fig:cube}, we establish a hierarchy of alignment objectives where universal values define ethical boundaries within which personal preferences are optimized, ensuring not compromising ethical principles.

% To address these challenges, we propose a comprehensive personalized alignment framework with three components: (1) preference memory management for maintaining user-specific knowledge, (2) personalized generation for response production incorporating personalized patterns, and (3) learning from feedback for improving preference memory and generation policies. While existing approaches treat universal values and personal preferences as parallel alignment objectives~\cite{zhang2024personalization}, we establish a hierarchy where universal values define ethical boundaries within which personal preferences are optimized, ensuring enhanced user satisfaction without compromising ethical principles, as illustrated in Figure~\ref{fig:cube}.
\begin{table*}[t]
\centering
\caption{A systematic comparison for personalization from traditional applications to LLMs. \textcolor{red}{Red} text highlights unique challenges and requirements introduced by personalized alignment in contrast to general alignment.}
\label{tab:personalization-comparison}
\begin{adjustbox}{max width=\linewidth}
\begin{tabular}{@{}p{3.6cm}|p{5.1cm}p{4.2cm}p{4.5cm}|p{3.2cm}p{5.8cm}@{}}
\hline
\multirow{1}{*}{\textbf{Aspects}} & \multirow{1}{*}{\textbf{Search}} & \multirow{1}{*}{\textbf{Advertising}} & \multirow{1}{*}{\textbf{Recommendation}} & \textbf{General Alignment} & \textbf{Personalized Alignment} \\
\hline
\textbf{User Preference Space} 
& Query History, Browsing 
& Demographics, Purchases
& Interaction Behavior
& \textcolor{gray}{\textit{Null}}
& \textcolor{red}{Personalized Preferences}\\
\hline
\textbf{Context Space} & Query, Location, Time & \textcolor{gray}{\textit{Null}} & \textcolor{gray}{\textit{Null}} & Dialogue Context & Dialogue Context\\
\hline
\textbf{Action Space}& Item Ranking & Ad Selection and Placement & Item Ranking & Textual Responses & Textual Responses\\
\hline
\textbf{Utility Function} & Information Accessibility\newline\cite{ndcg2017} & Revenue Generation\newline\cite{kumar2006scheduling} & Sustained User Engagement\newline\cite{choi2020online} & Universal Value\newline Alignment & Universal Value Alignment,\newline\textcolor{red}{Personalized Alignment} \\
\hline
\textbf{Application}& Google Search & Google Ads & TikTok, Netflix & ChatGPT & \textcolor{gray}{\textit{Growing Up ...}}\\
\hline
\textbf{Technical Challenges} & Query Understanding, Relevance-Personalization Trade-off & Budget Allocation & Cold Start, Preference Drift & Value Conflicts & Value Conflicts, \textcolor{red}{Preference Inference/Conflicts/Drift}, \textcolor{red}{Cold Start}, \textcolor{red}{Privacy}\\
\hline
\end{tabular}
\end{adjustbox}
\end{table*}

This survey presents the first systematic review of personalized alignment, with key contributions including: (1) A unified framework for personalized alignment; (2) A thorough analysis of existing methods and their synergies; and (3) An in-depth discussion of challenges and future directions. The survey is organized as follows: \S\ref{sec:background} introduces preliminaries, \S\ref{sec:framework} presents our framework, \S\ref{sec:techniques} analyzes existing techniques, \S\ref{sec:evaluation} and~\S\ref{sec:application} discuss evaluation and applications, \S\ref{sec:risk} and~\S\ref{sec:future} examine risks and future directions, and \S\ref{sec:conclusion} concludes with key insights.

% Prior surveys have extensively studied different aspects of LLMs and personalization. \citet{zhao2023survey} review the general development of LLMs, while \citet{ji2023ai} specifically examine AI alignment techniques. \citet{zhang2024personalization} investigate personalized applications. \citet{chen2024from} focus on role-playing LLMs that simulate specific personas. Distinctively, our survey specifically examines the integration of personalization into LLM alignment, providing a focused analysis of personalized generation and rewarding techniques.
Prior surveys have extensively studied different aspects of LLMs and personalization. \citet{zhao2023survey} review the general development of LLMs, while \citet{ji2023ai} specifically examine AI alignment techniques. \citet{zhang2024personalization} investigate personalized applications. \citet{chen2024from} focus on role-playing LLMs that simulate specific personas. While these works lay important foundations, personalized alignment presents fundamentally different challenges that have not been systematically addressed. Unlike traditional alignment that optimizes for universal values~\cite{ji2023ai}, personalized alignment must handle diverse objectives that vary across users and evolve over time while ensuring ethical boundaries. Unlike conventional personalization that focuses on surface-level preferences~\cite{zhang2024personalization}, our framework addresses deeper preference inference covering fundamental values and manages complex trade-offs between universal values and personal preferences. Distinctively, our survey provides the first comprehensive examination of these unique challenges, from privacy-preserving preference learning and real-time adaptation techniques to evaluation methods and risk mitigation strategies.

\section{Preliminaries}\label{sec:background}

% Analyzing user characteristics allows AI-driven tools to deliver highly personalized encounters that increase customer engagement. 
\subsection{AI Personalization}
AI personalization refers to tailoring AI systems to specific individual preferences~\cite{rossi1996value,montgomery2009prospects,wedel2016marketing}. Traditional personalization methods~\cite{linden2003amazon} have demonstrated significant success across search engines~\cite{pretschner1999ontology,speretta2005personalized}, advertising~\cite{zaichkowsky1994personal,tucker2014social}, and recommendation~\cite{resnick1997recommender,shani2011evaluating}. Formally, a personalized policy $\pi$ maps from the product space of user preferences and contextual factors to a feasible action: %. Specifically:
% , and possible actions to
%  probability distribution over actions
\begin{definition} Let $\mathcal{U}$ denote the preference space capturing user characteristics, $\mathcal{C}$ represent the context space encompassing temporal, spatial, and environmental variables, and $\mathcal{Y}$ define the action space containing all feasible system responses, then $\pi: \mathcal{U}\times\mathcal{C}\to\mathcal{Y}$. The optimal policy maximizes an application-specific utility function $R(\pi)$.
\end{definition}
This formalization manifests distinctly across applications, as illustrated in Table~\ref{tab:personalization-comparison}. While LLMs inherit similar personalization components, they face unprecedented challenges in personalizing generative processes while maintaining universal values.

\begin{figure*}[!t]
  \centering
  \includegraphics[width=\linewidth]{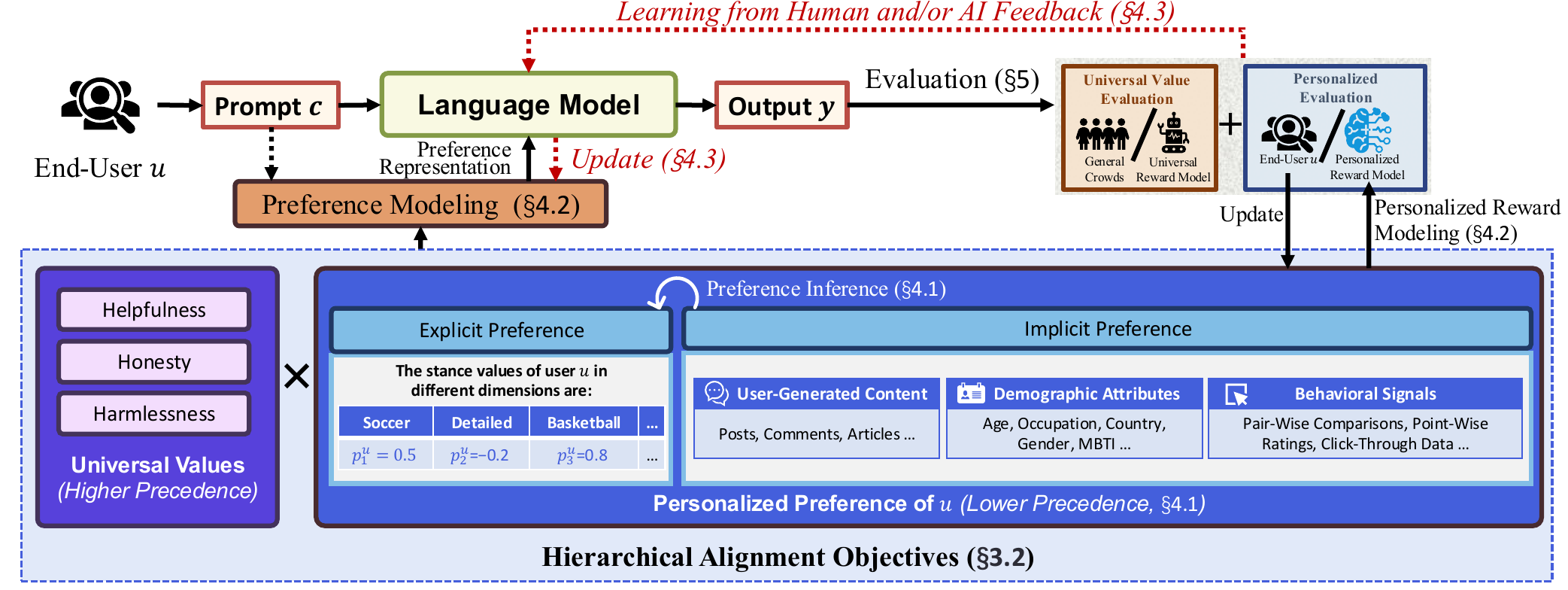}
  \caption{Overview of the personalized alignment framework.}
  \label{fig:framework}
  % \vspace{-3mm}
\end{figure*}

\begin{figure*}[!t]
\begin{tcolorbox}[
    colframe=pink!90!black,
    colback=pink!10!white,
    width=\textwidth,
    fontupper=\scriptsize,
    before skip=6pt,
    after skip=6pt,
]
\textbf{Task:}  Given the following preference memory, infer the user's preferences and generate preference-aligned responses.\\\\
\textbf{Preference Memory:}

    \begin{tabular}{@{}p{14.8cm}@{}}
    \toprule
    \textcolor{blue}{\texttt{\textbf{User-Generated Content:}}}\\
\textbf{Post (User):} ``Can you explain quantum computing?''\newline
\textbf{Response (LLM):} ``Quantum computing leverages quantum mechanical phenomena like superposition and entanglement. In quantum bits or qubits, states can exist simultaneously in multiple configurations, enabling parallel computations through quantum gates and ...''\newline
\textbf{Post (User):} ``That's too theoretical. Could you give real-world examples of how it's used in business or daily life? Theory without application isn't helpful.'' \textcolor{red}{(prefers practical applications)}\\
     \midrule
\textcolor{blue}{\texttt{\textbf{Demographic Attributes:}}} Professional Background: MBA with 5 years consulting experience; Learning Style: Action-oriented, prefers case studies over theoretical frameworks; Content Interaction: High engagement with business case analyses and implementation guides; Reading Pattern: Frequently accesses professional development and practical business strategy content \textcolor{red}{(prefers practical over theoretical content)}\\
\midrule
    \textcolor{blue}{\textbf{\texttt{Behavioral Signals:}}}\\
\textbf{Post (User):} ``I'm planning a trip to Japan next month. Any suggestions welcome!''\newline
    \textbf{Response A (LLM):} ``Here's a comprehensive cultural guide: Japan has 47 prefectures, 8 major regions, and a history dating back to 30,000 BCE ...''\newline
    \textbf{Response B (LLM):} ``Let's focus on practical tips: Learn basic greetings like 'arigatou' (thank you), avoid tipping as it's not customary, and remove shoes before entering homes ...''
    \newline
    \textbf{User's Preference:} Response B $\succ$ Response A 
    \textcolor{red}{(Prefers practical, accessible information)}\\\\
    
     \textbf{Post (User):} ``How can I improve my public speaking skills?''\newline \textbf{Response A (LLM):} ``Public speaking originated in ancient Greece with rhetoric principles developed by Aristotle. The fundamental elements include ethos, pathos, and logos...''\newline\textbf{Response B (LLM):} ``Here are specific techniques you can use: 1) Start with a relevant story, 2) Practice the 10-20-30 rule: 10 slides, 20 minutes, 30-point font...''\newline\textbf{User's Preference:} Response B $\succ$ Response A \textcolor{red}{(prefers actionable advice)}\\

% \midrule
% \textcolor{blue}{\texttt{\textbf{Explicit Preference:}}} Technical Content: $\downarrow$ \textcolor{red}{(Prefers simplified explanations)}, Practical Examples: $\uparrow$ \textcolor{red}{(Strongly favors real-world applications)}, 
% Detail Level: $\downarrow$ \textcolor{red}{(Prefers concise information)}\\
\bottomrule
    \end{tabular}
\newline\newline
\textbf{System's Preference Inference:} Based on the user's consistent preference for practical explanations over theoretical details across different domains, the user strongly prefers practical, accessible explanations with real-world applications over theoretical details\\
\newline
\textbf{Current Post (User):} ``What's the best way to understand blockchain?''\\
\textbf{Candidate Response 1 (LLM):} ``Blockchain is a distributed ledger technology utilizing cryptographic hashing functions and consensus mechanisms...''\\
\textbf{Candidate Response 2 (LLM):} ``Think of blockchain like a shared digital notebook - everyone has a copy, and when someone writes something new, everyone's copy gets updated automatically...''\\
\textbf{System's Alignment Objective:} Response 2 $\succ$ Response 1
\end{tcolorbox}
\caption{Example showing how a personalized alignment system infers user preferences from multiple information sources and generates preference-aligned responses.}
\label{example_preference_inference}
\end{figure*}

\subsection{The Development of LLMs}
The development of LLMs has progressed through three distinct stages, each characterized by different priorities in capability building. 
% and deployment strategies.

\paragraph{Pre-training Stage.} 
The initial stage focuses on developing foundation models through innovative architectures and training methods. The introduction of the Transformer architecture~\cite{vaswani2017attention} revolutionizes sequence modeling, while scaling laws~\cite{kaplan2020scaling} reveal the systematic relationships between model size, compute budget, and performance. Research on emergent abilities~\cite{wei2022emergent} demonstrates how certain capabilities only manifest beyond specific scale thresholds, such as few-shot learning~\cite{brown2020language} and chain-of-thought reasoning~\cite{wei2022chain}. These advances yield powerful but unaligned models, spanning the entire generation space (Figure~\ref{fig:cube}, entire space).
% While these advances built powerful models, they operated without explicit alignment considerations, resulting in models that could occupy any region in the universal value space (Figure~\ref{fig:cube}, entire space).

\paragraph{General Alignment Stage.} 

This stage further bridges the capability-usability gap through Supervised Fine-Tuning (SFT)~\cite{touvron2023llama} and Reinforcement Learning (RL)~\cite{christiano2017deep,ouyang2022training}. SFT optimizes models on human-curated examples, while RL utilizes human preferences for policy optimization. These methods achieve basic alignment with social preference but result in homogenized behavior (Figure~\ref{fig:cube}, \textcolor{red}{red} region), failing to accommodate individual differences~\cite{kirk2023understanding}.

% However, these methods, while successful in achieving basic alignment, result in homogenized behavior that represents an average across all users (Figure~\ref{fig:cube}, \textcolor{red}{red} region), failing to accommodate individual differences~\cite{kirk2023understanding,park2024rlhf}.

\paragraph{Deployment Stage.} 
Currently, this stage encompasses two main paradigms: API services like ChatGPT~\cite{Liu_2023} that adopt a one-size-fits-all approach, and task-specific agents~\cite{zhang2024generative,li2024personal} designed with specialized workflows. Both paradigms, however, face limitations in addressing diverse user needs. The key challenge lies in developing systematic personalization approaches that can adjust model behavior within universal value constraints to match individual user preferences (Figure~\ref{fig:cube}, \textcolor{blue}{blue} region), while maintaining operational efficiency.

\subsection{From Social to Personalized Preference}
\definecolor{mainblue}{RGB}{31,78,121}
\definecolor{secondblue}{RGB}{70,130,180}
\definecolor{lightbeige}{RGB}{245,245,220}
\tikzset{
    box/.style={
        draw,
        rounded corners,
        minimum height=3mm,
        text width=20mm,
        align=center,
        font=\scriptsize,
    },
    yellowbox/.style={box, 
        text width=110mm,
        fill=lightyellow},
    yellowlongbox/.style={box, 
        text width=130mm,
        fill=lightbeige, %lightyellow
        },
    bluebox/.style={box, 
        text width=20mm,
        fill=lightblue
        },
    greenbox/.style={box, fill=lightgreen},
    reference/.style={
        font=\scriptsize,
        text width=50mm,
        align=left
    },
    rotatedbox/.style={ % 专门为旋转文本设计的样式
        box,
        fill=lightblue, %secondblue,
        minimum width=3mm,  % 增加宽度以适应旋转
        minimum height=50mm, % 增加高度以适应旋转
        text width=3mm,
    },
    rotatedsubbox/.style={ % 专门为旋转文本设计的样式
        box,
        fill=secondblue, %lightblue,
        minimum width=3mm,  % 增加宽度以适应旋转
        minimum height=25mm, % 增加高度以适应旋转
        text width=3mm,
    }
}

\textit{``What should AI systems be aligned to?''} remains a longstanding research question. Traditional opinions, grounded in social choice theory~\cite{sen1986social}, advocate for aligning with aggregated human preferences~\cite{harsanyi1955cardinal,hare1981moral} to maximize collective utility. This has become the de facto standard in LLM development through voting-based preference aggregation across annotators~\cite{christiano2017deep,ouyang2022training}.
% harsanyi1975can
However, the approach faces fundamental challenges: preferences are often incomparable across different value systems~\cite{sen2017collective,korinek2022aligned,carrollai}; and centralizing alignment objectives risks imposing values of model creators onto all users~\cite{verdery2005socialism,scott2020seeing}.
% raising concerns about value pluralism and cultural diversity

These limitations motivate aligning LLMs with personalized preferences while maintaining universal ethical boundaries. This requires both direct user intervention during alignment~\cite{huang2024collective,guan2024amor} and consideration of broader ethical constraints~\cite{kalai1975other,oldenburg2024learning}. Such personalization relates to research on theory of mind~\cite{strachan2024testing} and emotional intelligence~\cite{sabour2024emobench} for understanding users' immediate mental states and emotions, while focusing on inferring preferences from their long-term characteristics and behaviors.
% Such personalization is fundamentally connected to theory of mind - enabling LLMs to understand users' beliefs, thoughts, and intentions~\cite{strachan2024testing}, as well as emotional intelligence - requiring models to recognize and respond appropriately to users' emotions~\cite{sabour2024emobench}. While these studies focus on users' immediate mental states and emotions, personalization focuses on inferring user preferences from their long-term characteristics and behavioral patterns.
% , and naturally accommodates value diversity. %while maintaining ethical boundaries. 
Formally, for any context $c\in\mathcal{C}$, the relationship between social and personalized preference can be expressed as:
\begin{equation*}
\pi_{\rm s}(y|c) = %\underset{u\in\mathcal{U}}
{\mathbb{E}_{u\in\mathcal{U}}}[\pi_{\rm p}(y|u,c)],
\end{equation*}
where $\pi_{\rm s}$ represents the social preference distribution adopted as the learning objective in current alignment practice, $\pi_{\rm p}$ denotes the personalized preference distribution for individual $u\in\mathcal{U}$. 
%and $h$ represents the universal human values.

% \gnote{they of mind (empathy)}

% gintis2010social

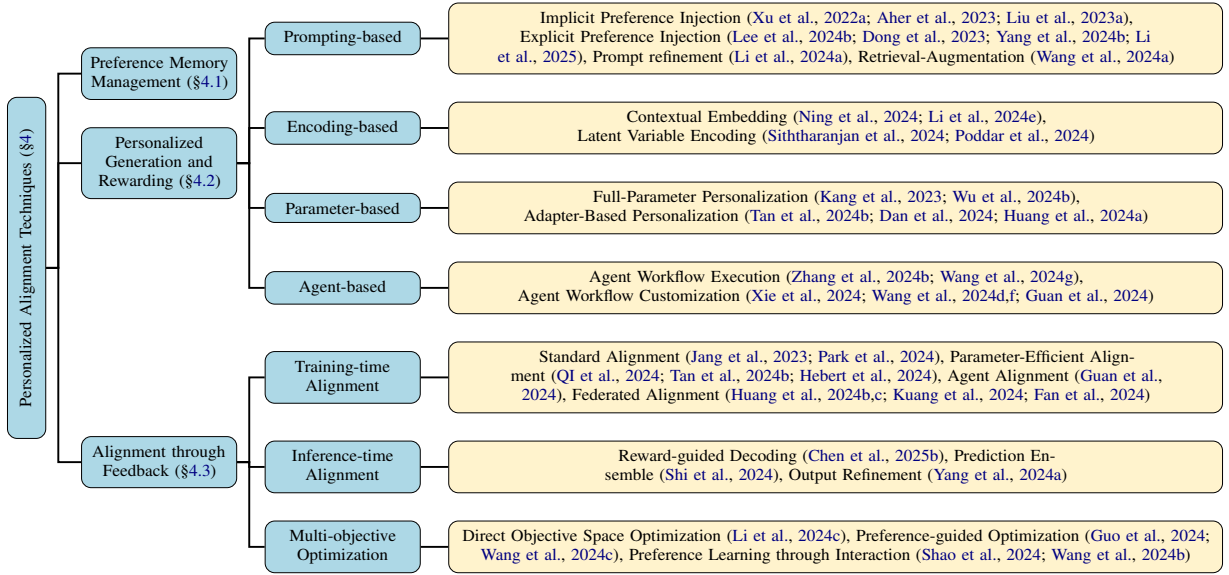
\begin{figure*}
\centering
\begin{adjustbox}{max width=\linewidth}
\begin{forest}
    for tree={
        grow'=east,
        parent anchor=east,
        child anchor=west,
        anchor=west,
        edge path={
            \noexpand\path[\forestoption{edge}]
            (!u.parent anchor) -- +(5pt,0) |- (.child anchor)\forestoption{edge label};
        },
        edge={thick},
        l sep=4mm,
        s sep=4mm
    }
    [\rotatebox{90}{{Personalized Alignment Techniques (\S\ref{sec:techniques})}}, rotatedbox
        [{Preference Memory Management~(\S\ref{subsec:Persona})},bluebox
        ]
        [{Personalized Generation and Rewarding~(\S\ref{subsec:personalize})},bluebox
            [Prompting-based, bluebox
                [{Implicit Preference Injection~\cite{xu2022beyond, aher2023using, liu2023chatgpt}, \\Explicit Preference Injection~\cite{lee2024aligning, dong2023steerlm, yang2024rewards,li20251000000usersuserscaling}, Prompt refinement~\cite{li2024learning}, Retrieval-Augmentation~\cite{wang2024unims}},yellowbox
                ]
            ]
            [Encoding-based, bluebox
                [{Contextual Embedding~\cite{ning2024user,li2024personalized}, \\Latent Variable Encoding~\cite{siththaranjan2024distributional,poddar2024personalizing}
                }, yellowbox
                ]
            ]
            [Parameter-based, bluebox
                [{Full-Parameter Personalization~\cite{kang2023llms, wu2024fine}, \\Adapter-Based Personalization~\cite{tan2024democratizing, dan2024p, huang2024selective}
                }, yellowbox
                ]
            ]
            [Agent-based, bluebox
                [{Agent Workflow Execution~\cite{zhang2024prospect, wang2024macrec}, \\Agent Workflow Customization~\cite{xie2024travelplanner, wang2024jumpstarter, wang2024surveyagent, guan2024amor}
                }, yellowbox
                ]
            ]
        ]
        [Alignment through Feedback~(\S\ref{subsec:align}),bluebox
            [Training-time Alignment,bluebox
                [{Standard Alignment~\cite{jang2023personalized,park2024rlhf}, Parameter-Efficient Alignment~\cite{qi2024fdlorapersonalizedfederatedlearning,tan2024democratizing,hebert2024persomapersonalizedsoftprompt}, Agent Alignment~\cite{guan2024amor}, Federated Alignment~\cite{huang2024collective,10571602,kuang2024federatedscope,fan2024fedrlhfconvergenceguaranteedfederatedframework}},yellowbox
                ]
            ]
            [Inference-time Alignment,bluebox
                [
                    {Reward-guided Decoding~\cite{chen2024pad}, Prediction Ensemble~\cite{shi2024decoding}, Output Refinement~\cite{yang2024metaaligner}}, yellowbox
                ]
            ]
            [Multi-objective Optimization,bluebox
                [
                    {Direct Objective Space Optimization~\cite{li2024differentiationmultiobjectivedatadrivendecision}, Preference-guided Optimization~\cite{guo-etal-2024-controllable,wang-etal-2024-conditional}, Preference Learning through Interaction~\cite{shao2024eliciting,10.1145/3627673.3679533}}, yellowbox
                ]
            ]
        ]
    ]
\end{forest}
\end{adjustbox}
\caption{A comprehensive taxonomy of personalized alignment techniques in LLMs.}
\label{fig:tax}
\end{figure*}

\section{Formalizing Personalized Alignment}\label{sec:framework}

Figure~\ref{fig:framework} illustrates our framework for personalized alignment~(\S\ref{cycle}), guided by hierarchical alignment objectives (\S\ref{objective}). 

\subsection{Personalized Alignment Cycle}\label{cycle}
Personalized alignment operates as a cycle with three key phases: \textbf{(1) Preference Memory Management:} maintaining user preference $u \in \mathcal{U}$~\cite{tam2006understanding,xu2022beyond}; \textbf{(2) Personalized Generation and Rewarding:} incorporating $u$ to capture preference patterns for generating or rewarding responses; \textbf{(3) Alignment through Feedback:} updating perference $u$ and the generation policy model through human or AI feedback.

\subsection{Alignment Objectives}\label{objective}
The objectives comprise two aspects: (1) universal human values (helpfulness, honesty, harmlessness)~\cite{askell2021general} as fundamental ethical constraints, and (2) personal preferences for user-specific needs. Given the potential conflicts and different priorities among these objectives~\cite{santurkar2023whose,rame2023rewarded}, universal values take precedence in defining ethical boundaries, within which personal preferences are optimized. Universal value weights are determined by ethical principles, while personal preference trade-offs are learned through user feedback.

\begin{figure*}[!t]
  \centering
  \includegraphics[width=\linewidth]{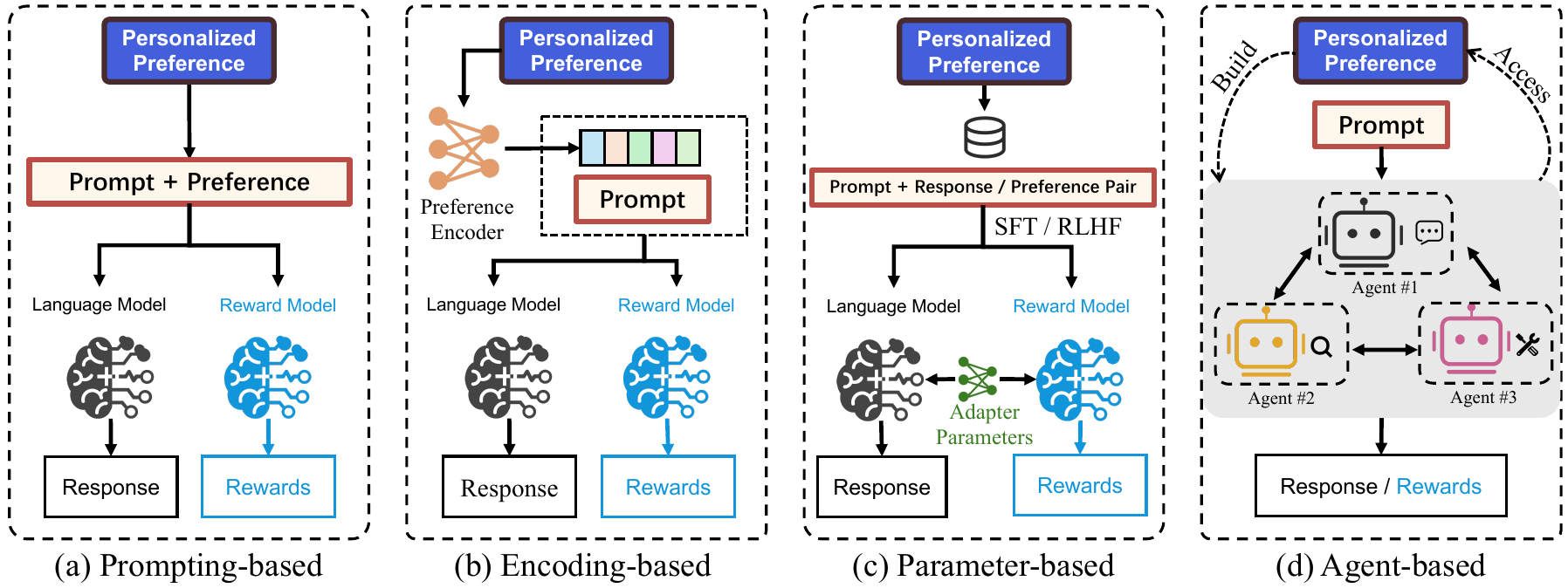}
  \caption{Approaches for personalized generation and rewarding, representing user preferences through (a) textual prompts, (b) encoded vectors, (c) trainable parameters, and (d) personalized workflows or accessible databases.}
  \label{fig:personalizatiob}
  % \vspace{-3mm}
\end{figure*}
\section{Techniques for Personalized Alignment}\label{sec:techniques}
This section examines three core components in the personalized alignment framework. Figure~\ref{fig:tax} illustrates the taxonomy of this section.

\subsection{Preference Memory Management}
\label{subsec:Persona}

Personalization requires a dedicated memory mechanism to utilize user-specific knowledge across interactions~\cite{10.1145/3604915.3608885}. In particular, the sparse distribution of user preferences in interactions poses significant challenges for retrieving relevant preferences from long contexts~\cite{zhao2025do,pan2025secom}. Since the capacity of LLMs in processing long-form text is beyond our scope, we assume the memory contains only preference-relevant information sources. This section examines both explicit and implicit preferences~\cite{jawaheer2014modeling,zhang2024personalization,li20251000000usersuserscaling}, and methods for inferring explicit preferences from various implicit sources.

\paragraph{Explicit Preferences.}

Explicit preferences represent directly stated user stances toward specific attributes or behaviors (i.e., preference dimensions). Formally, consider a preference space with $D$ dimensions $\{d_1,d_2,\cdots,d_D\}$. The explicit preference of a user $u\in\mathcal{U}$ can be represented by a normalized stance vector $\boldsymbol{p}^u=[p^{u}_1,p^{u}_2,\cdots,p^{u}_D]$. Each $p^{u}_i\in[-1,1]$ refers to the stance value for the dimension $d_i$, where positive values indicate favorable attitudes, negative values indicate unfavorable attitudes, and zero indicates neutrality.  The magnitude $|p^{u}_i|$ represents the strength of the stance. With these explicit preference representations, one can easily formalize how dimensional evaluations contribute to users' overall judgment of model responses, enabling straightforward personalized generation and reward computation. Specifically, given a model response $y\in\mathcal{Y}$ in context $c\in\mathcal{C}$, the response is evaluated with a user-agnostic rating vector $\boldsymbol{r}(y|c)=[r_1(y|c),r_2(y|c),\cdots,r_D(y|c)]$ in the preference space, where each dimension is rated independently of user preferences. The overall judgment of $u$ for $y$ is assumed as the dot product of the explicit preference vector and rating vector\footnote{The linear combination modeling is enough to capture complex preference patterns through appropriate dimension transformations (e.g., non-linear mappings). To improve interpretability and optimization efficiency, we suggest constructing orthogonal dimensions through variable substitution and linear transformation (e.g., Principal Component Analysis) when needed.}: 
$$
R^u(y|c)=(\boldsymbol{p}^u)^{\mathrm{T}}\boldsymbol{r}(y|c)%\sum_{i=1}^Dp^{u}_ir_i(c,y).
$$
This formulation enables direct optimization of personalized generation models by maximizing the expected overall judgment. Notably, the preference vector $\boldsymbol{p}^{u}$ may depend on both the context $c$ and dimensional ratings $\boldsymbol{r}$. For example, a user might prefer professional responses in technical discussions but favor empathetic ones in casual conversations. Furthermore, even in the same context, when a response receives high ratings in professionalism, the user's preference for response length might shift from negative (preferring concise) to positive (favoring detailed), demonstrating how ratings in one dimension can influence preferences in another. However, due to implementation complexity, most existing studies assume explicit preferences remain constant regardless of context and model responses~\cite{zhao2025do,li20251000000usersuserscaling}.
%\footnote{We adopt a deterministic rather than stochastic formulation of $p^u_i$ for practical implementation and interpretability.}. 
%formally expressed as $p^{u}_i=f^u(c,\{r_j(c,y)\}_{j=1}^D,d_i)$.

\paragraph{Implicit preferences.} Implicit preferences refer to indirect signals that reveal user characteristics, vaguely reflecting users' judgments on specific model responses. As exemplified in Figure~\ref{example_preference_inference}, these signals can be observed through: \textbf{(1) User-Generated Content:} Textual data such as social media posts, chat records, reviews, and articles that exhibit expertise levels, interests, and writing styles. \textbf{(2) Demographic Attributes:} Self-reported structured information, e.g., age, gender, educational background, etc. \textbf{(3) Behavioral Signals:} Interaction patterns captured through comparative judgments, ratings, click-through behaviors, etc. While these implicit signals do not directly state preferences, they offer rich contextual information for inferring user characteristics and tendencies without explicitly defining preference dimensions. However, unlike explicit preferences that enable direct computation of judgments for unseen responses through preference-rating aggregation, implicit preferences require data-driven learning to map indirect signals to personalization objectives, such as response generation and reward computation, necessitating careful design of preference modeling algorithms.

% In practical applications, explicit preferences are often unavailable or insufficient for comprehensive personalization.
\paragraph{Preference Inference.}
Explicit preferences offer precise personalization signals but remain sparse in practice due to users' reluctance, inability to articulate, or unawareness of their preferences~\cite{lee2024aligning}. Implicit preferences, while information-rich, present challenges in noise, redundancy, and inconsistency~\cite{preetha2014personalized}. Although end-to-end modeling from implicit preferences to personalized generation and rewarding is possible, such approaches often lack interpretability and control over the learned preference structure. These challenges motivate preference inference - a process that distills explicit preferences by projecting implicit preferences onto predefined structured preference spaces~\cite{lee2024aligning}. Given several examples reflecting implicit preferences of a user, the inference process typically involves three steps: (1) constructing a structured preference space by identifying key dimensions that capture user preferences; (2) estimating the user's explicit preferences on the preference space from individual examples, which are often partial, ambiguous, and noisy; and (3) aggregating potentially conflicting dimensional preferences across examples to obtain a robust unified explicit preference vector. Without considering contextual and rating effects, simple averaging of stance values suffices~\cite{li20251000000usersuserscaling}; otherwise, data-driven learning is needed to model how stance values vary with specific context and response rating vectors. Nevertheless, significant challenges remain in defining preference dimensions, as well as capturing the temporal evolution of user preferences~\cite{liu2015modeling}.

\subsection{Personalized Generation and Rewarding}
\label{subsec:personalize}
 % (wjf)
% \paragraph{persona selection}

% \paragraph{text-based representation and prompting-based generation/rewarding}

% \paragraph{latent variable-based representation and generation/rewarding}

% \paragraph{learnable parameter-based representation and generation/rewarding}

% \paragraph{agent-based generation/rewarding with customized workflow}

% There have been several efforts made to incorporate personal information into policy models for personalized text generation or to develop personalized reward models that guide the generation process. Depending on how personal information is integrated, these methods can be categorized into four groups: text-based and prompt-based methods, latent variable-based methods, learnable parameter-based methods, and agent-based methods.
Personalizing LLMs can be achieved by either directly personalizing generation policies or modulating generation distributions using personalized reward models. Both approaches employ four fundamental personalization mechanisms (as illustrated in Figure~\ref{fig:personalizatiob}): prompting-based methods injecting personal information into inputs, encoding-based methods encoding user features in intermediate representations, parameter-based methods adapting model parameters, and agent-based methods orchestrating personalized workflows.

% Adapting LLMs to individual preferences can be achieved by either directly personalizing generation policies or modulating generation distributions using personalized reward models. While targeting different learning components, both approaches share four fundamental mechanisms for incorporating personal information: prompting-based methods that explicitly inject personal information into model inputs, encoding-based methods that encode user features in intermediate representations, parameter-based methods that introduce user-specific adaptations to model parameters, and agent-based methods that orchestrate personalized workflows. 
% Despite their different technical foundations, all these approaches aim to bridge the gap between universal alignment and individual user preferences, enabling LLMs to maintain ethical boundaries while delivering personalized experiences.

% , e.g., dialogue history
% encompassing identity, personality traits, socio-cultural characteristics, and emotional preferences
 % derived such as browsing patterns and purchase records
\paragraph{Prompting-Based Personalization.} 
It augments prompts with personal information including implicit preferences (derived from user-generated content~\cite{xu2022long, li2024hello}, demographics~\cite{aher2023using, argyle2023out}, and behavioral signals~\cite{liu2023chatgpt, li2023text,bao2023tallrec, li2023preliminary}) and explicit preferences (specified dimensions, directions and weights~\cite{dong2023steerlm, lee2024aligning, yang2024rewards,li20251000000usersuserscaling}). Key challenges lie in managing extensive histories. Recent advances include prompt refinement~\cite{li2024learning} and retrieval-augmentation~\cite{wang2024unims}, allowing efficient information selection and presentation.

\paragraph{Encoding-based Personalization.} These methods represent preferences in latent spaces through: (1) contextual embeddings compressing user data into fixed representations~\cite{ning2024user,li2024personalized,shenfeld2025language}, and (2) latent variable frameworks learning preference distributions~\cite{siththaranjan2024distributional,poddar2024personalizing, gong2025latent,chen2025pal}, capturing uncertainty and multimodal patterns in user preferences. Both approaches balance preference preservation with computational efficiency.

\paragraph{Parameter-based Personalization.} 
These approaches encode user preferences directly into model parameters through (1) full-parameter personalization that modifies all parameters through fine-tuning~\cite{kang2023llms, li2023teach, wang2023rolellm} or reinforcement learning~\cite{jang2023personalized, wu2024fine} despite prohibitive computational costs for maintaining complete model copies per user, or (2) adapter-based personalization that introduces lightweight modules ($<1\%$ parameters) while keeping the base model frozen~\cite{tan2024personalized,tan2024democratizing, dan2024p, huang2024selective}. Key challenges include overfitting to limited personal data, catastrophic forgetting of general capabilities, and computational constraints.

\paragraph{Agent-based Personalization.} 
The approaches orchestrate LLM agents at two levels: workflow execution and customization. At the execution level, specialized modules (e.g., \textit{User}, \textit{Item}, \textit{Searcher}) in recommendation frameworks~\cite{zhang2024prospect, wang2024macrec, shu2024rah} collaborate to process personalization signals systematically. At the customization level, workflows dynamically adjust based on user preferences for domain-specific applications, like travel planning~\cite{xie2024travelplanner, chen2024travelagent, tang2024itinera, singh2024personal}, research support~\cite{wang2024surveyagent,zhangautonomous,openai2025introducing}, workplace assistance~\cite{wang2024jumpstarter}, and knowledge navigation~\cite{guan2024amor}. Key challenges include the intractability of optimizing agent workflows, autonomous preference integration across diverse scenarios, and managing substantial computational overhead.

\subsection{Alignment through Feedback}
\label{subsec:align}
This stage leverages user-specific rewards to align models with individual preferences, considering both training-time and inference-time approaches while balancing multiple alignment objectives. 
% universal values and personal preferences.

% Alignment through personalized feedback encompasses approaches that leverage user-specific rewarding signals to align model performance with personalized preference. This process involves two key aspects: when to perform alignment (training time vs. inference time), what information to learn from feedback, and how to align with multiple objectives: universal values that apply across all users and personalized preferences that vary among individuals.

\paragraph{Training-time Alignment.}
Training-time alignment strategies vary across personalization approaches and model accessibility. When base models are accessible, standard SFT and RL optimization can be directly applied~\cite{park2024rlhf,jang2023personalized} for prompting-based, encoding-based, and full-parameter personalization. For adapter-based and agent-based approaches, training focuses only on user-specific components: adapter parameters~\cite{qi2024fdlorapersonalizedfederatedlearning,tan2024democratizing,hebert2024persomapersonalizedsoftprompt} and agent modules~\cite{guan2024amor} respectively. When base models are inaccessible, personalized federated learning~\cite{10571602,kuang2024federatedscope,fan2024fedrlhfconvergenceguaranteedfederatedframework} enables privacy-preserving distributed training through architectural innovations~\cite{yi2024pfedmoedatalevelpersonalizationmixture,saadati2025pmixfedefficientpersonalizedfederated,tran2025privacypreserving} and dynamic adaptation~\cite{bao2023adaptivetesttimepersonalizationfederated,lee2024fedl2p}.

\paragraph{Inference-time Alignment.}
Recent advances achieve personalized alignment through inference-phase decoding modifications, avoiding costly training, including: (1) reward-guided decoding, where PAD~\cite{chen2024pad}, Amulet~\cite{zhang2025amulet} and CoS~\cite{he2025context} generates token-level personalized adjustments to steer model predictions; (2) prediction ensemble, where MOD~\cite{shi2024decoding} and Drift~\cite{kim2025drift} combine token-level predictions from multiple objective-specific models; and (3) output refinement, where MetaAligner~\cite{yang2024metaaligner} introduces an external correction model to refine outputs towards desired objectives. While these approaches offer immediate adaptability and training-free operation, they face inherent trade-offs between real-time adaptability and long-term preference learning due to the lack of parameter updates.

% Recent advances have introduced innovative approaches that achieve personalization purely through inference-phase modifications, primarily by redesigning the decoding process. These methods can be categorized based on their core mechanisms: (1) reward-guided decoding, where PAD~\cite{chen2024pad} generates token-level personalized rewards to steer the base model's predictions during inference; (2) prediction ensemble, where MOD~\cite{shi2024decoding} derives a theoretically-grounded way to optimally combine predictions from multiple objective-specific models through closed-form solutions; and (3) output refinement, where MetaAligner~\cite{yang2024metaaligner} introduces an external correction model to refine outputs from any policy model towards desired objectives, eliminating the need for model-specific training. These approaches share the advantages of training-free operation and immediate adaptability, making them particularly valuable for resource-constrained scenarios. However, they face inherent trade-offs between real-time adaptability and long-term preference learning, as the lack of parameter updates may limit their ability to capture complex, evolving user preferences.

\paragraph{Multi-objective Optimization.}
% Recent progress in multi-objective optimization has introduced sophisticated frameworks for personalized alignment to resolve conflicting objectives and user preferences, which can be divided into three categories based on training methodology: Multi-Objective End-to-End Learning~\cite{li2024differentiationmultiobjectivedatadrivendecision,10.1145/3627673.3679533}, Multi-Objective Reinforcement Learning~\cite{guo-etal-2024-controllable,wang-etal-2024-conditional} and Preference Elicitation through Feedback~\cite{shao2024eliciting}.\gnote{what is their links and differences?}  \textcolor{red}{These frameworks achieve multi-objective optimization by aligning with different personalized target sources, allowing for fine-grained control over various aspects of model behavior while ensuring alignment with personal preferences.}

% By integrating these various approaches to learning from feedback, LLMs can achieve more nuanced and effective personalization while maintaining their core capabilities and ethical principles. The choice between different strategies often depends on specific deployment constraints, such as computational resources, privacy requirements, and the need for real-time adaptation.
Three paradigms address the challenge of balancing inherent tensions among alignment objectives: (1) Direct objective space optimization~\cite{li2024differentiationmultiobjectivedatadrivendecision} uses specialized loss functions considering solution landscape and Pareto frontier, enabling explicit trade-offs; (2) Preference-guided optimization employs preference scores as control signals~\cite{guo-etal-2024-controllable} or conditional policies for dynamic adjustment~\cite{wang-etal-2024-conditional}, requiring careful preference specification; (3) Interactive preference learning leverages iterative feedback on policy pairs~\cite{shao2024eliciting} or hierarchical state tracking~\cite{10.1145/3627673.3679533}, adapting to users but demanding significant engagement. 
% Challenges remain in scaling to substantial objectives and learning robust preferences from limited feedback.

While sharing technical foundations with the traditional multi-objective optimization approaches, personalized alignment differs in its dynamic nature: it learns user-specific objectives and trade-offs through continuous interaction, rather than optimizing predefined objectives with fixed weights. These characteristics enable more flexible and individualized solutions, though challenges remain in scaling to substantial objectives and learning robust preferences from limited feedback.

\begin{table*}[!t]
    \centering
    \caption{Statistical details of the alignment dataset for training.}
    \label{tab:Dataset}
    \resizebox{\linewidth}{!}{\begin{tabular}{@{}lccccc@{}}
    \toprule
    \textbf{Dataset} & \textbf{Size} & \textbf{Feedback} (\textbf{Annotator}) & \textbf{Preference Dimensions} & \textbf{Preference Type} & \textbf{Task}\\
    \midrule
    \hline
    \rowcolor{gray!25} \multicolumn{6}{c}{\textbf{Universal Value Alignment}}\\
    \midrule
     
     \textbf{SHP}~\cite{ethayarajh2022understanding} & 385,000 & Rating (Human) & Helpfulness & Explicit Preference & Dialogue\\
     \hline
         \textbf{HH-RLHF}~\cite{bai2022training} & 169,550 & Ranking (Human) & Helpfulness, Harmlessness & Explicit Preference & Instruction-Following \\
    \hline

     \textbf{PRM800K}~\cite{lightman2023let} & 800,000 & Rating (Human) & Correctness & Explicit Preference & Math Problem-Solving\\
    \hline
    
    \textbf{HelpSteer}~\cite{wang2023helpsteer} & 37,000 & Rating (AI) & Helpfulness, Correctness, Coherence, Complexity, Verbosity & Behavioral Signals & Instruction-Following \\
    \hline
    
    \textbf{GPT-4-LLM}~\cite{peng2023instruction} & 52,000 & Rating (AI) & Helpfulness, Honesty, Harmlessness & Explicit Preference & Instruction-Following \\
    \hline
        \textbf{Nectar}~\cite{starling2023} & 182,954 & Ranking (AI) & Helpfulness, Harmlessness & Explicit Preference & Dialogue \\
    \hline

    \textbf{SafeRLHF}~\cite{dai2023safe} &1,000,000 & Ranking (Human) & Harmlessness & Explicit Preference & Instruction-Following \\
    \hline
    \textbf{OASST1}~\cite{kopf2024openassistant} & 161,443 & Rating (Human) & Helpfulness & Explicit Preference & Instruction-Following \\
    \hline
    \textbf{HelpSteer2}~\cite{wang2024helpsteer2} & 10,681 & Rating (AI) & Helpfulness, Correctness, Coherence, Complexity, Verbosity & Behavioral Signals & Instruction-Following \\
    
    \hline
    \textbf{Ultrafeedback}~\cite{cui2024ultrafeedback} & 63,967 & Rating (AI) & Helpfulness, Honesty, Instruction-Following, Truthfulness & Behavioral Signals & Instruction-Following \\
    \hline
    
    \textbf{Argilla-Distilabel-Capybara}& \multirow{2}{*}{7,560} & \multirow{2}{*}{Rating (AI)} & \multirow{2}{*}{Helpfulness, Honesty, Instruction-Following, Truthfulness} & \multirow{2}{*}{Behavioral Signals} & \multirow{2}{*}{Instruction-Following} \\
    \cite{distilabel-argilla-2024}\\
    \hline
    % \textbf{Distilabel-Capybara}~\cite{distilabel-argilla-2024} & 7,560 & Rating & AI & Helpfulness, Honesty, Instruction-Following, Truthfulness & Behavioral Signals & Instruction-Following \\
     
     \textbf{Beavertails}~\cite{ji2024beavertails} & 695,866 & Rating (Human) & Helpfulness, Harmlessness & Explicit Preference & Question-Answer, Expert Comparison\\
     \hline

     \multirow{2}{*}{\textbf{UltraInteract}~\cite{yuan2024advancing}} & \multirow{2}{*}{220,000} & \multirow{2}{*}{Ranking (AI)} & \multirow{2}{*}{Correctness} & \multirow{2}{*}{Explicit Preference} & Math Problem-Solving, Code \\
     &&&&&Generation, Logical Reasoning\\
     
    \midrule
    \hline
    \rowcolor{gray!25} 
    \multicolumn{6}{c}{\textbf{Personalized Alignment}}\\
    \midrule
    
    \textbf{Reddit TL;DR human feedback} & \multirow{2}{*}{123,169} & \multirow{2}{*}{Rating (Human)} & \multirow{2}{*}{\textcolor{gray}{\textit{Not Explicitly Defined}}} & \multirow{2}{*}{Behavioral Signals} & \multirow{2}{*}{Summarization} \\
    \cite{liu2020learning}\\
    \hline
    \textbf{WebGPT}~\cite{nakano2021webgpt} & 19,578 & Rating (Human) & \textcolor{gray}{\textit{Not Explicitly Defined}} & Behavioral Signals & Question-Answering \\
    \hline
    
\textbf{DSP}~\cite{cheng2023everyone} &13,000 & Text (AI) & Academy, Business, Entertainment, Literature\&Art & User-Generated Content & Instruction-Following \\
     \hline

     \textbf{Prometheus}~\cite{kim2023prometheus} & 100,000 & Rating (AI) & 1K fine-grained score rubrics & Behavioral Signals & Instruction-Following \\

    \hline    \textbf{PRISM}~\cite{kirk2024prism} & 8,010 & Rating (Human) & \textcolor{gray}{\textit{Not Explicitly Defined}} & Behavioral Signals & Instruction-Following \\
    \hline
    \multirow{2}{*}{\textbf{COMPO}}~\cite{kumar2024compo} & \multirow{2}{*}{1,389,750} & \multirow{2}{*}{Rating (Human)} & 187 community identifiers covering  & \multirow{2}{*}{Behavioral Signals} & \multirow{2}{*}{Dialogue} \\
    &&&science, finance, history, politics, gender/sexuality\\
    \hline
    
    \textbf{PersonalLLM}~\cite{2024personalllm} &10,402& Rating (AI) & \textcolor{gray}{\textit{Not Explicitly Defined}} & Behavioral Signals & Instruction-Following \\
    \hline
    
    \textbf{MULTIFACETED COLLECTION} & \multirow{2}{*}{197,000} & \multirow{2}{*}{Text (AI)} & \multirow{2}{*}{\textcolor{gray}{\textit{Not Explicitly Defined}}} & \multirow{2}{*}{User-Generated Content} & \multirow{2}{*}{Instruction-Following} \\
    \cite{lee2024aligning}\\
    \hline
    \textbf{CodeUltraFeedback} & \multirow{2}{*}{10,000} & \multirow{2}{*}{Rating (AI)} & Instruction-Following, Code Explanation, Coding
Style, & \multirow{2}{*}{Behavioral Signals} & \multirow{2}{*}{Code Generation} \\
\cite{weyssow2024codeultrafeedback}&&& Code Complexity and Efficiency, Code Readability\\
    \hline
    \multirow{2}{*}{\textbf{\textsc{AlignX}}~\cite{li20251000000usersuserscaling}}&\multirow{2}{*}{1,311,622}&\multirow{2}{*}{Ranking (AI)} & 90 dimensions covering fundamental human needs, &Explicit Preference,&\multirow{2}{*}{Instruction-Following}\\
    &&&universal human values, and prevalent interest tags&Implicit Preference&\\
     
    \bottomrule
    \end{tabular}}
\end{table*}

\subsection{Resources}
\label{subsec:resource}
High-quality preference datasets are fundamental to training personalized generation and reward models. We present a comprehensive overview of alignment training datasets in Table~\ref{tab:Dataset}.
\begin{table*}[t]
\centering
\caption{A systematic categorization of evaluation methods for personalized alignment.}
\label{tab:metrics}
\begin{adjustbox}{max width=\linewidth}
\begin{tabular}{@{}p{3.5cm}ccccp{6cm}p{7.5cm}@{}}
\toprule
\multirow{2}{*}{\textbf{Evaluation Method}} & {\textbf{Reference}} & {\textbf{Reference}} & \textbf{Personalized} & \textbf{Evaluation}& \multirow{2}{*}{\textbf{Description}} & \multirow{2}{*}{\textbf{Weakness}}\\
&\textbf{Response} & \textbf{Response Pair}& \textbf{Evaluation Model}&\textbf{Mode}&&\\
\midrule
\textbf{Win Rate}\newline\cite{khanov2024args} & \xmark & \xmark & \cmark&Pair-wise & Using personalized LLMs to determine the better one of two generations & Computationally expensive; Requiring substantially powerful LLMs\\
\hline
\textbf{Rating}\newline\cite{chen2024pad} & \xmark & \xmark & \cmark &Point-wise& Using personalized reward models to directly rate the generation & High sensitivity to prompt variations; Requiring significant data for model training\\
\hline
\textbf{Alignment Accuracy}\newline\cite{park2024rlhf} & \xmark & \cmark & \xmark&Point-wise & Measuring correct preference ordering of reference response pairs  & Limited to evaluating white-box models that provide probability access; Requiring many high-quality reference pairs\\
\hline
\textbf{BLEU}\newline\cite{papineni2002bleu} & \cmark & \xmark & \xmark&Point-wise & $N$-gram precision between the generation and reference response & Struggling to capture preferences; Ineffective for open-ended responses~\cite{guan-huang-2020-union}\\
% \hline
% \textbf{ROUGE}\newline\cite{lin2004automatic} & \cmark & \xmark & \xmark&Point-wise & Recall-oriented comparison between the generation and reference response & \textit{ditto} \\
\hline\hline
\textbf{Human Evaluation}&\xmark&\xmark&\xmark&Point-wise& Assessment through human raters & High subjectivity and variance; Poor scalability\\

\bottomrule
\end{tabular}
\end{adjustbox}
\end{table*}

\begin{table*}[!t]
    \centering
    \caption{Summary of alignment benchmarks. \textbf{Win}: {Win Rate}; \textbf{ACC}: {Alignment Accuracy}.}
    \label{tab:benchmark}
    \begin{adjustbox}{max width=\linewidth}
    \begin{tabular}{@{}lcccccc@{}}
    \toprule
    \textbf{Benchmark} & \textbf{Size} & \textbf{Feedback (Annotator)} & \textbf{Preference Dimensions}& \textbf{Evaluation Metrics}  & \textbf{Preference Type} & \textbf{Task}\\
    
    \midrule
    \rowcolor{gray!25} \multicolumn{7}{c}{\textbf{Universal Value Alignment}}\\
    \midrule
    \textbf{TruthfulQA}~\cite{lin2021truthfulqa} & 817  & Ranking (Human) & Truthfulness& Rating/Win/ACC/BLEU/ROUGE& Explicit Preference & Instruction-Following  \\
    \hline
    
    \textbf{HHH-Alignment}~\cite{askell2021general} & 221 & Ranking (Human)& Helpfulness, Honesty, Harmlessness  & ACC& Explicit Preference & Instruction-Following  \\

    \hline
    
    \textbf{Self-Instruct}~\cite{wang2022self} & 11,800& Ranking (Human\&AI) & Helpfulness & Rating/Win/BLEU/ROUGE & User-Generated Content& Instruction-Following  \\

    \hline
    \textbf{MT Bench}~\cite{zheng2023judging} & 3,360 & Ranking (Human\&AI)& Helpfulness & Rating/Win/ACC/BLEU/ROUGE& Explicit Preference & Instruction-Following  \\
    \hline
    
        \textbf{Advbench}~\cite{zou2023universal} & 150 & Ranking (AI)  & Harmlessness & Rating/Win/BLEU/ROUGE& User-Generated Content& Instruction-Following  \\
    \hline

    \textbf{AlpacaEval 2.0}~\cite{dubois2024length} & 805& \textcolor{gray}{\textit{No Feedback}} & Helpfulness  & Rating/Win/BLEU/ROUGE& User-Generated Content & Instruction-Following  \\
    \hline

    \textbf{Arena Hard}~\cite{li2024crowdsourced} & 500& \textcolor{gray}{\textit{No Feedback}} & Helpfulness  & Rating/Win/ACC/BLEU/ROUGE & Explicit Preference& Instruction-Following, Code Generation  \\
    \hline

    \textbf{RewardBench}~\cite{lambert2024rewardbench} & 2,985 & Ranking (Human) & Helpfulness  & ACC & Explicit Preference& Chat, Chat-Hard, Safety, Reasoning  \\

    \midrule
    \hline
    \rowcolor{gray!25} \multicolumn{7}{c}{\textbf{Personalized Alignment}}\\
    \midrule

    \textbf{P-Soups}~\cite{jang2023personalized} & 50 & \textcolor{gray}{\textit{No Feedback}}  & Expertise, Informativeness, Style& Rating/Win/BLEU/ROUGE & User-Generated Content& Instruction-Following  \\
    \hline
    
    \textbf{LaMP}~\cite{salemi2023lamp} & 25,095 &\textcolor{gray}{\textit{No Feedback}}& \textcolor{gray}{\textit{Not Explicitly Defined}}& Rating/Win/BLEU/ROUGE & User-Generated Content & Text Classification, Text Generation \\
    \hline

    \textbf{OpinionQA}~\cite{santurkar2023whose} & 1,176& Ranking (Human) & \textcolor{gray}{\textit{Not Explicitly Defined}}  & ACC & Behavioral Signals& Question-Answering  \\
    \hline

    \textbf{GlobalOpinionQA}~\cite{durmus2023towards} & 2,556 & Ranking (Human) & \textcolor{gray}{\textit{Not Explicitly Defined}} & ACC& Behavioral Signals & Question-Answering  \\
    \hline
    \textbf{FLASK}~\cite{ye2023flask} & 1,740  & \textcolor{gray}{\textit{No Feedback}} & \textcolor{gray}{\textit{Not Explicitly Defined}}& Rating/Win/BLEU/ROUGE & User-Generated Content & Instruction-Following  \\
    \hline
    \textbf{REGEN}~\cite{sayana2024retrievalgeneratingnarrativesconversational} & 403,000 & Rating (Human) & \textcolor{gray}{\textit{Not Explicitly Defined}} & BLEU/ROUGE/Similarity Scores & User-Generated Content, Behavioral Signals & Conversational Recommendations\\
    \hline
    
    \multirow{2}{*}{\textbf{LongLaMP}~\cite{kumar2024longlampbenchmarkpersonalizedlongform}}&\multirow{2}{*}{9,658}&\multirow{2}{*}{\textcolor{gray}{\textit{No Feedback}}}&\multirow{2}{*}{\textcolor{gray}{\textit{Not Explicitly Defined}}}&\multirow{2}{*}{ROUGE}&{User-Generated Content,}&Email Completion, Abstract Generation,\\
    &&&&&Demographic Attributes&Review Writing, Topic Writing\\
    \hline
    \textbf{PGraphRAG}~\cite{au2025personalizedgraphbasedretrievallarge}& 10,000 & Text, Rating (Human) & \textcolor{gray}{\textit{Not Explicitly Defined}} & Rating/Win/BLEU/ROUGE & User-Generated Content, Behavioral Signals & Long and Short Text
Generation, Classification\\
    \hline
    \textbf{PersonalLLM}~\cite{zollo2025personalllm}&10,000&\textcolor{gray}{\textit{No Feedback}}&\textcolor{gray}{\textit{Not Explicitly Defined}}&Rating/Win&Behavioral Signals&Instruction-Following\\
    \hline
    \textbf{ALOE}~\cite{wu-etal-2025-aligning}&100&\textcolor{gray}{\textit{No Feedback}}&\textcolor{gray}{\textit{Not Explicitly Defined}}&{Win}&Demographic Attributes&Instruction-Following\\
    \hline
    \textbf{PERSONA}~\cite{castricato-etal-2025-persona}&3,868&Ranking~(AI)&\textcolor{gray}{\textit{Not Explicitly Defined}}&Win/ACC&Demographic Attributes&Instruction-Following\\
    \hline
    \multirow{2}{*}{\textbf{\textsc{PREFEVAL}}~\cite{zhao2025do}}&\multirow{2}{*}{3,000}&\multirow{2}{*}{Ranking (Human)}&\multirow{2}{*}{\textcolor{gray}{\textit{Not Explicitly Defined}}}&\multirow{2}{*}{Rating/ACC}&Explicit Preference, User-Generated&\multirow{2}{*}{Instruction-Following}\\
    &&&&&Content, Behavioral Signals\\
    \bottomrule
    \end{tabular}
    \end{adjustbox}
\end{table*}

\section{Evaluation of Personalized Alignment}\label{sec:evaluation}
Evaluating personalized alignment presents unique challenges beyond traditional LLM evaluation, requiring simultaneous assessment of universal value alignment and individual preference satisfaction. This section examines current evaluation approaches and their limitations.
\subsection{Metrics}
Table~\ref{tab:metrics} summarizes existing metrics for personalized alignment. Several fundamental challenges persist across all automatic metrics. Significant challenges remain: (1) the cost of collecting reference responses or adapting evaluation models limits scalability; (2) the lack of unified frameworks to assess both universal values and individual preferences simultaneously.

\subsection{Benchmarks}
% While universal value alignment has established benchmarks, personalized alignment benchmarks remain limited, as indicated in Table~\ref{tab:benchmark}. Critical challenges include: (1) insufficient scale in dataset size and preference diversity; (2) lack of protocols for evaluating cross-user generalization and robustness to preference shifts.

While universal value alignment has established benchmarks, personalized alignment benchmarks remain limited, as indicated in Table~\ref{tab:benchmark}. Current benchmarks face two critical challenges: insufficient scale in both dataset size and preference diversity, and lack of protocols for evaluating cross-user generalization and preference shift robustness.

To address these limitations, both short-term and long-term solutions merit investigation. In the short term, structured protocols for multi-faceted preference collection~\cite{li20251000000usersuserscaling} and longitudinal studies for preference evolution can enhance existing benchmarks. Long-term directions include developing human-in-the-loop metrics~\cite{WU2022364} and privacy-preserving evaluation mechanisms. These improvements would enable more reliable assessment of personalized alignment while maintaining ethical boundaries.

\section{Application of Personalized Alignment}\label{sec:application}
% (wjf)
Personalized LLMs extend beyond chatbots to transform productivity, daily experience, and social welfare through user-adaptive interactions.
\subsection{Personal Assistants}
Personalized LLMs advance traditional AI assistance (e.g., open-domain QA~\cite{high2012era}) by providing domain-specific support, including: intelligent coding companions with tailored completions and optimizations~\cite{dai2024mpcoder,hiraki2024personalization,nejjar2025llms,koohestani2025rethinking}, research aids generating customized literature reviews~\cite{wang2024surveyagent,lin2024paper}, and workplace assistants for personalized task management~\cite{zhang2024tablellm,wang2024officebench,wang2024jumpstarter,teufelberger2024llm}.

% wu2019session
\subsection{Consumer Applications}
Personalized LLMs enhance digital services via user-generated content analysis. In recommendations, they transcend traditional approaches~\cite{su2009survey,he2017neural} via instruction-based generation~\cite{liu2023chatgpt,li2023text, li2023preliminary} or multi-agent frameworks~\cite{wang2024macrec}. Entertainment applications include adaptive companions~\cite{zhou2023characterglm}, role-playing~\cite{chen2024from}, and interactive narratives~\cite{wu2024role,sun2025drama}, though raising ethical concerns regarding emotional manipulation.

\subsection{Public Services}
 % transform public services through adaptive natural language interactions.
In education, Personalized LLMs enable individualized learning experiences~\cite{yadav2023contextualizing,kabir2023llm} and support educators with automated analytics and material generation~\cite{jeon2023large, leiker2023prototyping, koraishi2023teaching, kasneci2023chatgpt}. In healthcare, they advance patient care~\cite{cascella2023evaluating, gebreab2024llm} by integrating medical knowledge with individual profiles~\cite{abbasian2023conversational,jo2023understanding,sallam2023chatgpt}, though requiring rigorous validation~\cite{mirzaei2024clinician}.

\section{Risks of Personalized Alignment}
\label{sec:risk}

% Personalized alignment introduces interconnected risks at both individual and societal levels~\cite{kirk2024benefits}. At the individual level, key risks include: (1) portrait abuse, where malicious actors exploit user profiles for targeted attacks~\cite{al2024open,wang2024deeplearningmodelsecurity,sabour2025humandecisionmakingsusceptibleaidriven}; (2) information leakage through model inversion and membership inference~\cite{huang-etal-2022-large,li2024model}; and (3) bias reinforcement across gender, political, and geographical dimensions~\cite{raza2024fair,kotek2023gender,motoki2024more}, potentially creating amplifying feedback loops~\cite{he2024cos}.

% These individual risks cascade to societal challenges: (1) access disparities due to technical literacy and resource constraints~\cite{wilson2003social,sanders2021digital,lythreatis2022digital}; and (2) social polarization~\cite{weidinger2022taxonomy,shelby2022identifying} through selective exposure and information asymmetry~\cite{abdelzaher2020paradoxinformationaccessgrowing,gurkan2024personalizedcontentmoderationemergent}.

Personalized alignment introduces interconnected risks at both individual and societal levels~\cite{kirk2024benefits}. At the individual level, key risks include: (1) portrait abuse, where malicious actors exploit user profiles for targeted attacks~\cite{al2024open,wang2024deeplearningmodelsecurity,sabour2025humandecisionmakingsusceptibleaidriven} - for example, crafting personalized phishing messages based on inferred user preferences or manipulating responses to exploit known behavioral patterns; (2) information leakage through model inversion and membership inference~\cite{huang-etal-2022-large,li2024model}, which could reveal sensitive user preferences and interaction history; and (3) bias reinforcement across gender, political, and geographical dimensions~\cite{raza2024fair,kotek2023gender,motoki2024more}, potentially creating amplifying feedback loops~\cite{he2024cos} where the system increasingly reinforces existing biases through personalized responses.

These individual risks cascade to broader societal challenges: (1) access disparities due to technical literacy and resource constraints~\cite{wilson2003social,sanders2021digital,lythreatis2022digital}, as sophisticated personalization may primarily benefit tech-savvy users while marginalizing others; and (2) social polarization~\cite{weidinger2022taxonomy,shelby2022identifying} through selective exposure and information asymmetry~\cite{abdelzaher2020paradoxinformationaccessgrowing,gurkan2024personalizedcontentmoderationemergent}, where personalized content delivery can create echo chambers and fragment public discourse.

To mitigate these risks, both technical and operational measures are crucial. Technical solutions include federated preference learning to protect user privacy, differential privacy mechanisms for preference data, and robust debiasing techniques during model training. Operational guidelines should encompass regular bias auditing, transparent preference management policies, and mechanisms for users to understand and control their preference profiles. However, implementing these safeguards while maintaining personalization effectiveness remains an ongoing challenge.

\section{Key Challenges and Future Directions}\label{sec:future}
Current challenges in personalized alignment span three aspects: foundational methodologies, technical implementation, and practical deployment.

Foundational challenges focus on understanding and modeling user preferences: (1) capturing complex personalized preferences that evolve temporally~\cite{chandrashekaran1996modeling}, are influenced by social context~\cite{izuma2013social}, vary across scenarios (e.g., work vs. life situations)~\cite{smailagic2002application}, and exhibit response-dependent trade-offs between different dimensions (e.g., quality vs. speed)~\cite{li2024differentiationmultiobjectivedatadrivendecision}; (2) addressing data-related challenges, including both the scarcity of high-quality personalization data (in terms of scale, diversity, and temporal dynamics) and the cold-start problem where systems lack initial user data for meaningful personalization~\cite{schein2002methods}; (3) developing evaluation frameworks that can reliably assess personalization quality beyond surface-level metrics; (4) ensuring models understand user needs and follow user preference~\cite{zhang2023instruction, zhao2025do}; (5) enabling proactive preference exploration and learning, where systems actively engage with users to discover and refine preference inference, and balance exploration of new preferences with exploitation of known preferences.

Technical challenges concern the implementation of personalized systems: (1) integrating multimodal signals (e.g., text, vision, audio) to better understand and generate personalized content~\cite{pi2024strengthening,pi2025personalized}; (2) incorporating user preferences into long-chain reasoning while preserving interpretability and contextual consistency~\cite{river2025personalization}; (3) balancing performance and computational efficiency for real-time personalization; (4) ensuring personalized models can effectively generalize from simple to complex tasks with weak supervisions~\cite{burns2023weak,kenton2024on}.

Practical challenges address deployment concerns while prioritizing universal values: (1) protecting user privacy and security while maintaining personalization effectiveness; (2) ensuring fairness across different user groups, system transparency, and positive societal impact; (3) coordinating multiple personalized agents while preventing harmful emergent behaviors; (4) safely removing dangerous capabilities while preserving personalization performance and ensuring system accessibility.

\section{Conclusion}\label{sec:conclusion}
We present a comprehensive examination of personalized alignment in LLMs, demonstrating its critical role in bridging universal value alignment with individual user needs. Our unified framework, that encompasses preference memory management, personalized generation and rewarding, and alignment through feedback, provides a structured approach for advancing this field. By analyzing various implementation strategies and diverse applications, we reveal the significant potential and current limitations of personalized alignment. 

While substantial challenges remain, they also indicate promising research directions. Success in personalized alignment requires continued innovation in technical approaches, robust evaluation frameworks, and careful consideration of ethical implications. As LLMs become prevalent, personalized alignment will be crucial for serving diverse users effectively.

\section{Limitations}

While this survey strives to provide a comprehensive overview of personalized alignment in LLMs, several limitations should be acknowledged:

First, given the rapid development of the field, some very recent advances may not be included. Additionally, due to space constraints, we could not exhaustively cover all existing techniques and applications.

Second, our categorization of approaches into preference memory management, personalized generation and rewarding, and alignment through feedback, while useful for organization, may oversimplify the complex interrelationships between these components.

Third, the evaluation methods and metrics discussed in this survey largely reflect current practices, which may not fully capture the nuanced aspects of personalization quality. The field lacks standardized evaluation frameworks, making it challenging to compare different approaches objectively.

Finally, while we attempted to include diverse perspectives on ethical considerations and societal impacts, our discussion may not comprehensively address all potential implications of personalized alignment, particularly in emerging application scenarios.

\bibliography{custom}
\end{document}